\documentclass[sn-mathphys-ay]{sn-jnl}% Math and Physical Sciences Author Year Reference Style
\usepackage{graphicx}%
\usepackage{multirow}%
\usepackage{amsmath,amssymb,amsfonts}%
\usepackage{amsthm}%
\usepackage{mathrsfs}%
\usepackage[title]{appendix}%
\usepackage{xcolor}%
\usepackage{textcomp}%
\usepackage{manyfoot}%
\usepackage{booktabs}%
\usepackage{algorithm}%
\usepackage{algorithmicx}%
\usepackage{algpseudocode}%
\usepackage{listings}%
\theoremstyle{thmstyleone}%
\newtheorem{theorem}{Theorem}%  meant for continuous numbers
\newtheorem{lemma}[theorem]{Lemma}
\newtheorem{corollary}[theorem]{Corollary}

\theoremstyle{thmstyletwo}%
\newtheorem{remark}{Remark}%
\newtheorem{assume}{Assumption}[section]

\theoremstyle{thmstylethree}%
\newtheorem{definition}{Definition}%

\begin{document}
	
	\title[Concentration properties of fractional posterior in 1-bit matrix completion]{Concentration properties of fractional posterior in 1-bit matrix completion}
	
	
	\author*{\fnm{The Tien} \sur{Mai}}\email{the.t.mai@ntnu.no}
		
	\affil{
		\orgdiv{Department of Mathematical Sciences}, 
		\orgname{Norwegian University of Science and Technology}, 
		\orgaddress{%\street{Street}, 
			\city{Trondheim}, \postcode{7034}, %\state{State},
			\country{Norway}}
		}

\abstract{
The problem of estimating a matrix based on a set of its observed entries is commonly referred to as the matrix completion problem. In this work, we specifically address the scenario of binary observations, often termed as 1-bit matrix completion. While numerous studies have explored Bayesian and frequentist methods for real-value matrix completion, there has been a lack of theoretical exploration regarding Bayesian approaches in 1-bit matrix completion. We tackle this gap by considering a general, non-uniform sampling scheme and providing theoretical assurances on the efficacy of the fractional posterior. Our contributions include obtaining concentration results for the fractional posterior and demonstrating its effectiveness in recovering the underlying parameter matrix. We accomplish this using two distinct types of prior distributions: low-rank factorization priors and a spectral scaled Student prior, with the latter requiring fewer assumptions. Importantly, our results exhibit an adaptive nature by not mandating prior knowledge of the rank of the parameter matrix. Our findings are comparable to those found in the frequentist literature, yet demand fewer restrictive assumptions. 
}




\maketitle

\section{Introduction}
Matrix completion has been extensively explored in the fields of machine learning and statistics, attracting considerable attention in recent years due to its relevance to various contemporary applications such as recommendation systems \citep{bobadilla2013recommender,koren2009matrix}, including the notable Netflix challenge \citep{bennett2007netflix}, image processing \citep{ji2010robust,han2014linear}, genotype imputation \citep{chi2013genotype,jiang2016sparrec}, and quantum statistics \citep{gross2011recovering}. Although completing a matrix in general is often deemed infeasible, seminal works by \cite{CandesT10,candes2010matrix,CandesRecht} have demonstrated its potential feasibility under the assumption of a low-rank structure. This assumption aligns naturally with practical scenarios, particularly in recommendation systems, where it implies the presence of a limited number of latent features that capture user preferences. Various theoretical and computational approaches to matrix completion have been proposed and investigated, as in \cite{tsybakov2011nuclear,lim2007variational,salakhutdinov2008bayesian,recht2013parallel,chatterjee2015matrix,mai2015,alquier2020concentration,chen2019inference}.

The previously mentioned studies primarily focused on matrices with real-numbered elements. However, in many practical situations, the observed elements are often binary, taking values from the set $\{-1, 1\}$. This type of data is prevalent in diverse contexts, such as voting or rating data, where responses typically involve binary distinctions like ``yes/no", ``like/dislike", or ``true/false". Tackling the challenge of reconstructing a matrix from incomplete binary observations, known as 1-bit matrix completion, was initially investigated in \cite{Davenport14}. Subsequent studies in this field have been conducted by various researchers \citep{cai2013,klopp2015adaptive,hsieh2015pu,cottet20181bit,herbster2016mistake,alquier2019estimation}, most of whom have taken a frequentist approach. However, there remains a gap in the literature concerning the theoretical assessment of Bayesian methodologies in this domain.

In this study, we aim to address this gap by focusing on a generalized Bayesian approach, where we utilize a fractional power of the likelihood. This leads to what is commonly referred to as fractional posteriors or tempered posteriors, as elucidated in \cite{bhattacharya2016bayesian, alquier2020concentration}. It is noteworthy to emphasize that generalized Bayesian methods, where the likelihood is substituted by its fractional power or by a concept of risk, has garnered increased attention in recent years, as demonstrated by various works such as \cite{hammer2023approximate,jewson2022general,yonekura2023adaptation,mai2017pseudo, matsubara2022robust,medina2022robustness,grunwald2017inconsistency,bissiri2013general,yang2020alpha,lyddon2019general,syring2019calibrating,Knoblauch,mai2023reduced,hong2020model}. Additionally, we tackle this problem by considering a general, non-uniform sampling scheme. While a general sampling scheme for 1-bit matrix completion has been examined in \cite{klopp2015adaptive}, our requirements are less stringent than theirs.

Initially, we present results concerning the employment of a widely used low-rank factorized prior distribution. Such priors have demonstrated practical efficacy, as evidenced in works such as \cite{cottet20181bit, lim2007variational, salakhutdinov2008bayesian}. However, due to the typically large dimensionalities of matrix completion problems, employing low-rank factorized priors necessitates intricate Markov Chain Monte Carlo (MCMC) adaptations, which can be computationally expensive and lack scalability. Consequently, in practical applications, variational inference is often favored for such priors, as discussed in works like \cite{cottet20181bit, lim2007variational, babacan2012sparse}.
We present novel results regarding the consistency and concentration properties of the fractional posterior. Specifically, we establish concentration results for the recovering distribution within the $\alpha$-R\'enyi divergence framework. Consequently, as particular instances, we derive concentration outcomes relative to metrics such as the Hellinger metric. Furthermore, we broaden our investigation to establish concentration rates for parameter estimation utilizing specific distance measures such as the Frobenius norm. Our findings are comparable to those in the frequentist literature as documented in \cite{Davenport14, cai2013}, and \cite{klopp2015adaptive}.

In addition to the aforementioned type of prior, we also undertake theoretical examination utilizing a spectral scaled Student prior. This prior, introduced by \cite{dalalyan2020exponential}, shares conceptual similarities with a hierarchical prior discussed in \cite{yang2018fast}. The spectral scaled Student prior enables posterior sampling through Langevin Monte Carlo, a gradient-based sampling technique that has recently garnered considerable attention in various high-dimensional problems, as observed in \cite{durmus2017nonasymptotic, durmus2019high, dalalyan2017theoretical, dalalyan2019user}. We demonstrate that by employing this prior, it is possible to achieve concentration results for the fractional posterior without necessitating a boundedness assumption, as is typically required for low-rank factorization priors.

The remainder of the paper is structured as follows. In Section \ref{sec:Model}, we introduce the notations essential for our work and discuss the problem of 1-bit matrix completion. We also present the fractional posterior along with the low-rank factorization prior in this section. Section \ref{sc_main1} presents the results pertaining to the low-rank factorization prior, while Section \ref{sc_main2} is dedicated to the outcomes obtained using the spectral scaled Student prior. All technical proofs are consolidated in Section \ref{sc_proofs}.

\section{Notations and method}
\label{sec:Model}

\subsection{Notations}

For any integer $m$, let $[m]=\{1,\dots,m\} $. Given integers $m$ and $k$, and a matrix $M \in \mathbb{R}^{m\times k}$, we write $\Vert M \Vert_\infty: = \max_{(i,j)\in [m]\times [k]} |M_{ij}| $. For a matrix $ M $, its spectral norm is denoted by $ \Vert M \Vert $, its Fobenius norm is denoted by $ \Vert M \Vert_F = \sqrt{\sum_{ij}M^2_{ij}} $, and its nuclear norm is denoted by $ \Vert M \Vert_* $ (the sum of the singular values). 

Let $\alpha\in(0,1)$, the $\alpha$-R\'enyi divergence  between two probability distributions $Q $ and $R$ is  defined by
\begin{align*}
D_{\alpha}(Q ,R)  
=
\frac{1}{\alpha-1} \log \int \left(\frac{{\rm d}Q }{{\rm d}\mu}\right)^\alpha \left(\frac{{\rm d}R}{{\rm d}\mu}\right)^{1-\alpha} {\rm d}\mu  
,
\end{align*}
where  $\mu$ is any measure such that $Q \ll \mu$ and $R\ll \mu$. The Kullback-Leibler (KL) divergence is defined by
\begin{align*}
\mathcal{K}(Q ,R)  
= 
\int \log \left(\frac{{\rm d}Q }{{\rm d}R} \right){\rm d}Q  \text{ if } Q  \ll R
\text{, }
+ \infty \text{ otherwise}.
\end{align*}

\subsection{1-bit matrix completion}

We assume that the observed data $ (w_1,Y_1), \ldots, (w_n,Y_n) $ are \text{i.i.d.} (independent and identically distributed) random variables drawn from a joint distribution characterized by a matrix $M^* \in \mathbb{R}^{d_1\times d_2}$, denoted by $ P_{M^*} $. Additionally, we assume that $ (\omega_s)_{s=1}^n \in ([d_1]\times [d_2])^n $ are \text{i.i.d.} and denoted by $\Pi$ its marginal distribution. These indices correspond to observations, denoted by $ (Y_s)_{s=1}^n \in ({-1,+1})^n $, distributed accordingly:
\begin{equation}
\label{eq_logisticmodel}
Y| \omega
=
\begin{cases}
1 & \text{ with probability } f(M^*_\omega),
\\
-1 & \text{ with probability } 1- f(M^*_\omega),
\end{cases} 
\end{equation}
where $f $ is the logistic link function
$ f(x) = \frac{\exp(x)}{1+\exp(x)}$. This model is similar to \cite{klopp2015adaptive}.
In this model, we have the likelihood of the observations as $ L_n (M):= \prod_{s=1}^{n}f(M_{\omega_i})^{1_{\left[Y_i =1 \right] }} (1-f(M_{\omega_i}))^{1_{\left[Y_i =-1 \right]}} $.

In this study, we operate under the assumption that the rank of \( M^* \), denoted as \( r \), is substantially smaller than its dimensions, specifically \( r \ll \min(d_1, d_2) \). This is a prevalent assumption in 1-bit matrix completion research, \citep{Davenport14,cai2013,cottet20181bit,klopp2015adaptive,alquier2019estimation}.

We concentrate on the fractional posterior for \( \alpha \in (0,1) \), as discussed in \cite{bhattacharya2016bayesian,alquier2020concentration}, which is formulated as follows:
\begin{align*}
\pi_{n,\alpha}(M)
\propto
L_n^\alpha (M) \pi(M)
.
\end{align*}
In the case $ \alpha =1 $, one recovers the traditional posterior distribution.

We define the mean estimator as
\begin{equation}
\label{eq_mean_estimator}
\hat{M} 
:=
\int M \pi_{n,\alpha}({\rm d}M)  
.
\end{equation}

\subsection*{Low-rank factorization prior}

In Bayesian matrix completion methodologies \citep{babacan2012sparse,salakhutdinov2008bayesian,lim2007variational}, a prevalent concept involves decomposing a matrix into two matrices in order to establish a prior distribution on low-rank matrices. It is commonly acknowledged that any matrix with a rank of $r$ can be decomposed as follows:
$
M=LR^\top, \, 
L\in \mathbb{R}^{d_1 \times r}, \, 
R \in \mathbb{R}^{d_2 \times r}.
$
This approach is grounded in the assumption that the underlying matrix $M^*$ exhibits a low rank, or is at least well approximated by a low-rank matrix.

However, in practical scenarios, the rank of the matrix is typically unknown. Thus, for a fixed $ K \in \{1, \ldots, \min(d_1,d_2) \} $, one can express $M=LR^\top$ with $L\in \mathbb{R}^{d_1 \times K}$, $R \in \mathbb{R}^{d_2 \times K}$. Subsequently, potential ranks $r\in[K]$ are adjusted by diminishing certain columns of $L$ and $R$ to zero. To address this, the reference \cite{cottet20181bit} considers the following hierarchical model:
\begin{align*}
\gamma_k &\stackrel{iid}\sim \pi^\gamma,
\, \forall k \in [K] ,
\\
L_{i,\cdot},R_{j,\cdot}|\gamma 
&\stackrel{iid}\sim \mathcal{N}(0,\textrm{diag}(\gamma)),
\,\, \forall (i,j) \in [d_1] \times [d_2].
\end{align*}
The prior distribution on the variances $\pi^\gamma$ plays a crucial role in controlling the shrinkage of the columns of $L$ and $R$ towards zero. It is common for $\pi^\gamma$ to follow an inverse-Gamma distribution \citep{salakhutdinov2008bayesian}. This hierarchical prior distribution bears resemblance to the Bayesian Lasso proposed in \cite{park2008bayesian}, and particularly resembles the Bayesian Group Lasso \citep{kyung2010penalized}, where the variance term follows a Gamma distribution.

The paper \cite{cottet20181bit} shows that the Gamma distribution is also a possible alternative in matrix completion,
both for theoretical results and practical considerations.
Thus all the results in this paper are stated under the assumption that
$\pi^\gamma$ is either the
Gamma or the inverse-Gamma distribution: $\pi^\gamma = \Gamma(a,b) $, or $\pi^\gamma = \Gamma^{-1}(a,b)$.
In this study, we regard $ a $ as a fixed constant, while $ b $ is seen as a small parameter requiring adjustment.

\section{Main results}
\label{sc_main1}
For $r\geq 1$ and $B>0$, we define $\mathcal{M}(r,B)$ as the set of pairs of matrices $(\bar{U},\bar{V})$, with dimensions $ d_1 \times K$ and $ d_2\times K$ respectively, satisfying that: $\|\bar{U}\|_{\infty} \leq B$, $\|\bar{V}\|_{\infty} \leq B$ and $\bar{U}_{i,\ell}=0$ for $i>r$ and  $\bar{V}_{j,\ell}=0$ for $j>r$. Similar to \cite{cottet20181bit,alquier2020concentration}, we make the following assumption on the true parameter matrix.

\begin{assume}
	\label{assume_factorbounded}
	We assume that $M^* = \bar{U}\bar{V}^t$ for $(\bar{U},\bar{V})\in \mathcal{M}(r,B)$.
\end{assume}

The following theorem presents the first consistency result for the fractional posterior in 1-bit matrix completion with Gaussian priors which frequently employed in practical applications.

\begin{theorem}
	\label{theorem_main}
	Assume that Assumption \ref{assume_factorbounded} holds. Then, there is a small enough $ b >0 $ such that  
	\begin{equation*}
	\mathbb{E} \left[ \int D_{\alpha}( P_{M} , P_{M^*}) \pi_{n,\alpha}({\rm d} M ) \right]
	\leq 
	\frac{1+\alpha}{1-\alpha}\varepsilon_n
	.
	\end{equation*}
	where
	$$\varepsilon_n = C_{a,B}\frac{ r(d_1+d_2)\log(nd_1d_2)}{n} 
	,
	$$
	for some universal constant $C_{a,B} $ depending only on $ a, B $. Specifically, the result remains valid for the selection $ b= B^2/[512(nd_1d_2)^4 K^2 \max^2(d_1,d_2)] $.
\end{theorem}

It is reminded that all technical proofs are postponed to Section \ref{sc_proofs}. The main argument is based on a general scheme for fractional posteriors derived in \cite{bhattacharya2016bayesian,alquier2020concentration}.

In practical applications, it is noted that $ b= B^2/[512(nd_1d_2)^4 K^2 \max^2(d_1,d_2)] $ may not the best choice; rather, \cite{alquier2014bayesian,alquier2020concentration} suggests employing cross-validation to select \( b \). Ensuring a small \( b \) is crucial in practical situations to guarantee a reliable approximation of low-rank matrices \citep{alquier2014bayesian,alquier2020concentration}.

The following theorem introduces the first concentration results for the fractional posterior in 1-bit matrix completion when utilizing commonly employed Gaussian priors in practical applications.

\begin{theorem}
	\label{theorem_main2}
	Assume that Assumption \ref{assume_factorbounded} holds. Then,
	for a sufficiently small \( b > 0 \), such as \( b = \frac{B^2}{512(nd_1d_2)^4 K^2 \max^2(d_1,d_2)} \), it holds that
	$$
	\mathbb{P}\left[
	\int \mathcal{D}_{\alpha}(P_{M} , P_{M^*}) \pi_{n,\alpha}({\rm d}M )  
	\leq
	\frac{2(\alpha+1)}{1-\alpha} \varepsilon_n
	\right] \geq 1-\frac{2}{n\varepsilon_n}
	$$
	where,
	$$
	\varepsilon_n = C_{a,B}\frac{ r(d_1+d_2)\log(nd_1d_2)}{n} 
	,
	$$
	for some universal constant $C_{a,B} $ depending only on $ a$ and $ B $.
\end{theorem}

\begin{remark}
	It is important to note that our results are formulated without prior knowledge of $ r $, the rank of the true underlying parameter matrix. This aspect highlights the adaptive nature of our results, indicating their ability to adjust and perform effectively regardless of the specific rank of the true underlying parameter matrix.
\end{remark}

Put
\begin{equation*}
c_\alpha 
=
\begin{cases}
\frac{2(\alpha+1)}{1-\alpha}, \alpha \in [0.5,1) ,
\\
\frac{2(\alpha+1)}{\alpha}, \alpha \in (0, 0.5)
.
\end{cases}
\end{equation*}
\begin{corollary}
	\label{cor_concentration_Hellinger}
	As a special case, Theorem \ref{theorem_main} leads to a concentration result in terms of the classical Hellinger distance
	\begin{equation}
	\label{eq_Hellinger_mainresults}
	\mathbb{P}\left[
	\int H^2(P_{M} , P_{M^*} ) \pi_{n,\alpha}({\rm d} M )  
	\leq 
	c_\alpha 
	\varepsilon_n \right]
	\geq 
	1-\frac{2}{n\varepsilon_n}
	.
	\end{equation}
\end{corollary}

\begin{remark}
	The rate specified in \eqref{eq_Hellinger_mainresults}, of the order $ r(d_1+d_2)\log(n)/n $, bears resemblance to that observed in prior studies in frequentist literature like \cite{klopp2015adaptive} when examining a general sampling framework. To elaborate further, Theorem 1 and Lemma 9 in \cite{klopp2015adaptive} delve into the recovery of the distribution \( f(M) \); however, they necessitate stricter assumptions. In comparison, our findings in \eqref{eq_Hellinger_mainresults} demonstrate a faster rate than those outlined in \cite{Davenport14} where their results is of order $ \sqrt{r(d_1+d_2)\log(\max(d_1,d_2))/n} $.
\end{remark}

To derive results directly concerning the parameter matrix, we must make use of the following assumption.

\begin{assume}
	\label{assum_observation_prob}
	We assume that there exist a constant $ C_1 >0 $, such that,
	\begin{equation*}
	\min_{ i\in[d_1] , j\in[d_2]} 
	\mathbb{P} (\omega_i =(i,j) )
	\geq C_1
	.
	\end{equation*}
\end{assume}

This assumption guarantees that each coefficient has a non-zero probability of being observed. For instance, with the uniform distribution, we can express it as $ C_1 = 1/(d_1d_2) $. This assumption was initially introduced in \cite{klopp2014noisy} within the classical unquantized (continuous) matrix completion setting. It is also used for 1-bit matrix completion under a general sampling distribution, as demonstrated in \cite{klopp2015adaptive}.

\begin{assume}
	\label{assum_boundedness}
	We assume that $ \|M^*\|_\infty \leq \kappa < \infty $ and there exist a constant $ C_\kappa >0 $ such that 
	\begin{equation*}
	C_\kappa 
	=
	\inf_{|x|\leq \kappa}
	\frac{f'(x)^2}{8f(x)(1-f(x))}
	,
	\end{equation*}
	where $ f(x) = e^x/(1+e^x) $.
\end{assume}
Assumption \ref{assum_boundedness} stands as a cornerstone requirement essential for deriving insights into estimation errors. It was first introduced in \cite{Davenport14} and has since served as a fundamental premise in various prior works, including \cite{cai2013} and \cite{klopp2015adaptive}.

We are ready to state our main results regarding the recovering of the parameter matrix.

\begin{theorem}
	\label{theorem_estimation}
	Under the same assumption as in Theorem \ref{theorem_main2} and additionally assuming that Assumption \ref{assum_observation_prob} and Assumption \ref{assum_boundedness} hold. We have that
	\begin{equation}
	\label{eq_estimation_concentration}
	\mathbb{P}\left[
	\int  \frac{ \| M - M^* \|^2_F
	}{d_1 d_2} 
	\pi_{n,\alpha}({\rm d} M )  
	\leq 
	\frac{c_\alpha}{C_1C_\kappa}
	\varepsilon_n \right]
	\geq 
	1-\frac{2}{n\varepsilon_n}
	.
	\end{equation}	
	and 
	\begin{equation}
	\label{eq_bound_meanestimator}
	\mathbb{P}\left[
	\frac{ \| \hat{M} - M^* \|^2_F
	}{d_1 d_2} 
	\leq 
	\frac{c_\alpha}{C_1C_\kappa}
	\varepsilon_n \right]
	\geq 
	1-\frac{2}{n\varepsilon_n}
	.
	\end{equation}
\end{theorem}

\begin{remark}
	Up to a logarithmic factor, the error rate for the mean estimator in the squared Frobenius norm, given in \eqref{eq_bound_meanestimator}, is of order $ r(d_1+d_2)/n $ which is minimax-optimal according to Theorem 3 in \cite{klopp2015adaptive}. The result stated in \eqref{eq_bound_meanestimator} is achieved by applying Jensen's inequality to the mean. By employing similar methods, one can readily derive outcomes for other estimator derived from the fractional posterior, such as the median, drawing upon insights provided in \cite{merkle2005jensen}.
\end{remark}

\begin{remark}
	Under the uniform sampling assumption and that $ \| X\|_{\infty}\leq \gamma $,  Theorem 1 in \cite{Davenport14} presented results of order
	$
	\sqrt{r(d_1+d_2)/n } 
	$. 
	A similar
	result using max-norm minimization was also obtained in \cite{cai2013}. The paper \cite{klopp2015adaptive} proves a faster estimation error rate as  
	$
	r(d_1+d_2)\log(d_1+d_2)/n 
	$.
	A comparable result to \cite{klopp2015adaptive} is also established in \cite{alquier2019estimation}. Subsequently, this rate has been recently enhanced to \( r(d_1+d_2)/n \), without the presence of a logarithmic term, in \cite{alaya2019collective} (refer to Theorem 7).  Consequently, the work presented in \cite{alaya2019collective} attains the precise minimax estimation rate of convergence for 1-bit matrix completion.
\end{remark}

\begin{remark}
	It is noteworthy that our findings are established within a general sampling framework. In contrast to the requirements set forth in \cite{klopp2015adaptive}, our approach necessitates only that the probability of observing any entries is strictly positive, without imposing additional assumptions. This aspect further enhances the robustness of employing a fractional posterior.
\end{remark}

\section{Results with a spectral scaled Student prior}
\label{sc_main2}

We have opted to initially present results in Section \ref{sc_main1} with factorization-type priors, as they are widely favored in the matrix completion literature for utilization with MCMC or Variational Bayes (VB) methods. However, another spectral scaled Student prior has garnered particular interest due to its promising outcomes, whether employed with VB \citep{yang2018fast} or Langevin Monte Carlo, a gradient-based sampling method \citep{dalalyan2020exponential}. This prior has previously been applied in different problems involving matrix parameters \citep{mai2023bilinear,mai2023reduced}.

With $\tau>0$, we consider the following spectral scaled Student prior, given as
\begin{align}
\label{prior_scaled_Student}
\pi_{st} (M)
\propto
\det (\tau^2 \mathbf{I}_{d_1} + MM^\intercal )^{-(d_1+d_2+2)/2}.
\end{align}

\noindent This prior possesses the capability to introduce approximate low-rankness in matrices \( M \). This is evident from the fact that
\( \pi_{st} (M)
\propto
\prod_{j=1}^{d_1} (\tau^2 + s_j(M)^2 )^{- (d_1+d_2+2)/2 },
\)
where \( s_j(M) \) represents the \( j^{th} \) largest singular value of \( M \). Consequently, the distribution follows a scaled Student's t-distribution evaluated at \( s_j(M) \), which induces approximate sparsity on \( s_j(M) \), as discussed in \cite{dalalyan2012sparse,dalalyan2012mirror}. Thus, under this prior distribution, the majority of \( s_j(M) \) tend to be close to \( 0 \), suggesting that \( M \) is approximately low-rank.

We are now present a consistency result using the spectral scaled Student prior.

\begin{theorem}
	\label{theorem_main_prior2}
	For $ \tau = 1/n $, we have that
	$$
	\mathbb{E}\left[
	\int \mathcal{D}_{\alpha}(P_M,P_{M^*})  \pi_{n,\alpha}({\rm d} M ) \right]
	\leq 
	\frac{1+\alpha}{1-\alpha}\varepsilon_n
	$$
	where
	$$
	\varepsilon_n 
	= 
	\frac{
		2 r (d_1 + d_2 +2) \log \left( 1+ \frac{ n \| M^* \|_F}{ \sqrt{2r }} \right) 
	}{n}
	.
	$$
\end{theorem}

The proofs of this section can be found in Section \ref{sc_proofs2}. It is noted that in the rate $ \varepsilon_n $ outlined in Theorem \ref{theorem_main_prior2} and Theorem \ref{theorem_main_prior2_prob} below, the condition $ r = {\rm rank} (M^* ) \neq 0 $ is not necessary. This is because we interpret $ 0\log(1+0/0) $ as $ 0 $ for the scenario where $ r^* = 0 $ and $ M^* = 0 $.

The next Theorem presents a concentration result for the fractional posterior.

\begin{theorem}
	\label{theorem_main_prior2_prob}
	For $ \tau = 1/n $, we have that
	$$
	\mathbb{P}\left[
	\int \mathcal{D}_{\alpha}(P_M,P_{M^*}) \pi_{n,\alpha}({\rm d}M )  \leq
	\frac{2(\alpha+1)}{1-\alpha} \varepsilon_n
	\right] \geq 1-\frac{2}{n\varepsilon_n}
	$$
	where
	$$
	\varepsilon_n 
	= 
	\frac{
		2 r (d_1 + d_2 +2) \log \left( 1+ \frac{ n \| M^* \|_F}{ \sqrt{2r }} \right) 
	}{n}
	.
	$$
\end{theorem}

\begin{remark}
	We do not assert that $ \tau = 1/n $, in both Theorem \ref{theorem_main_prior2} and \ref{theorem_main_prior2_prob}, represents the optimal selection. In practical applications, users can utilize cross-validation to fine-tune the value of $ \tau $.
\end{remark}

\begin{remark}
	It is interesting to observe that by utilizing the spectral scaled Student prior described in \eqref{prior_scaled_Student}, we are not required to impose a boundedness assumption on $ M^* $, as was necessary in the previous section with low-rank factorized priors. Furthermore, the additional logarithmic factor in Theorem \ref{theorem_main_prior2} and Theorem \ref{theorem_main_prior2_prob} can be further simplified. This can be achieved by employing the inequality $ \| M^* \|_F \leq \| M^* \| \sqrt{r} $, resulting in  $ \log (1+ n\| M^* \|) $.
\end{remark}

Similar to Theorem \ref{theorem_estimation}, with the inclusion of additional assumptions, we can derive concentration results for recovering the underlying matrix parameter as well as results for the mean estimator defined in \eqref{eq_mean_estimator}.
\begin{theorem}
	\label{theorem_estimation2}
	Under the same assumption as in Theorem \ref{theorem_main_prior2_prob} and additional assume that Assumption \ref{assum_observation_prob} and Assumption \ref{assum_boundedness} hold. We have that
	\begin{equation}
	\label{eq_estimation_concentration2}
	\mathbb{P}\left[
	\int  \frac{ \| M - M^* \|^2_F
	}{d_1 d_2} 
	\pi_{n,\alpha}({\rm d} M )  
	\leq 
	\frac{c_\alpha}{C_1C_\kappa}
	\varepsilon_n \right]
	\geq 
	1-\frac{2}{n\varepsilon_n}
	.
	\end{equation}	
	and 
	\begin{equation}
	\label{eq_bound_meanestimator2}
	\mathbb{P}\left[
	\frac{ \| \hat{M} - M^* \|^2_F
	}{d_1 d_2} 
	\leq 
	\frac{c_\alpha}{C_1C_\kappa}
	\varepsilon_n \right]
	\geq 
	1-\frac{2}{n\varepsilon_n}
	.
	\end{equation}
\end{theorem}

\begin{remark}
	Similar to the outcomes detailed in Section \ref{sc_main1}, the results presented in this section for the spectral scaled Student prior do not necessitate prior knowledge of $ r $, the rank of the true underlying parameter matrix. This underscores the adaptive nature of our results, demonstrating their capacity to adjust and perform effectively, regardless of the rank of the true underlying parameter matrix.
\end{remark}

\section{Proofs}
\label{sc_proofs}

\subsection{Proofs for Section \ref{sc_main1}}

\begin{proof}[\textit{\textbf{Proof of Theorem} \ref{theorem_main}}]
	As the logistic loss is $ 1 $-Lipschitz, the log-likelihood satisfies that 
	\begin{align*}
	\left\vert\log f(x)-\log f(y) \right\vert 
	\leq 
	|x-y|
	.
	\end{align*}
	One has that
	\begin{align*}
	\mathcal{K} (P_{M^*}, P_M) 
	&	\leq
	\frac{1}{d_1  d_2} \sum_{i\in[d_1]}
	\sum_{j\in[d_2]} \Pi_{ij}
	|M^*_{ij} - M_{ij} |
	\\
	&	\leq
	\frac{1}{d_1  d_2} \sum_{i\in[d_1]}
	\sum_{j\in[d_2]} 
	|M^*_{ij} - M_{ij} |
	\\
	&	\leq
	\frac{\| M^* -M \|_F }{ \sqrt{ d_1 d_2} }
	,
	\end{align*}
	where $ \Pi_{ij}\leq 1 $ is the probability to observe the $ (i,j) $-th entry. For any $(U,V)$ in the support of $\rho_n$, given in \eqref{eq_specific_rhon}, one has that
	\begin{align*}
	\|M^* - UV^t\|_F 
	& = 
	\| \bar{U}\bar{V}^t-\bar{U}V^t+\bar{U}V^t-UV^t\|_F \\
	& \leq  \|\bar{U}(\bar{V}^t-V^t)\|_F+\|(\bar{U}-U)V^t\|_F \\
	& \leq  \|\bar{U}\|_F \|\bar{V}-V\|_F+\|\bar{U}-U\|_F \|V^t\|_F \\
	& \leq  
	d_1 d_2 \|\bar{U}\|_{\infty}^{1/2} \|\bar{V}-V\|_\infty^{1/2}
	+d_1 d_2 \|V\|_{\infty}^{1/2} \|\bar{U}-U\|_\infty^{1/2} \\
	& \leq 
	d_1 d_2 \delta^{1/2}  
	[B^{1/2} + (B+\delta)^{1/2}  ] 
	\\
	& \leq 
	2 d_1 d_2 \delta^{1/2} (B+\delta)^{1/2} 
	\\
	& \leq 
	2^{3/2} d_1 d_2 \delta^{1/2} B^{1/2}  
	.
	\end{align*}
	Therefore,
	\begin{align}
	\mathcal{K} (P_{M^*}, P_M) 
	\leq
	\frac{\| M^* -M \|_F }{ \sqrt{ d_1 d_2} }
	\leq
	\sqrt{ \delta 2^{3} d_1 d_2  B}
	\label{eq_klbound_1}
	.
	\end{align}
	For $\delta=B/[8(nd_1 d_2)^2]$ that satisfies $0<\delta<B$, we have that
	\begin{align}
	\label{eq_bound_Frobenius}
	\|M^* - UV^t\|_F 
	\leq
	B/n
	\end{align}
	and
	\begin{align*}
	\int \mathcal{K} (P_{M^*}, P_M) \rho_n(dM)
	\leq
	\frac{B}{n\sqrt{d_1d_2}}
	.
	\end{align*}
	Now, from Lemma \ref{lm_boundKL}, we have that
	\begin{equation*}
	\frac{1}{n}\mathcal{K}(\rho_n,\pi) 
	\leq 
	\frac{
		2(1+2a) r(d_1 + d_2) \left[ \log(nd_1 d_2)
		+
		C_a
		\right] }{n}
	.
	\end{equation*}	
	We now can apply Theorem 2.6 in \cite{alquier2020concentration} with $ \rho_n $ in \eqref{eq_specific_rhon} and
	\begin{equation*}
	\varepsilon_n = C_{a,B}\frac{ r(d_1+d_2)\log(nd_1d_2)}{n} 	
	\end{equation*}
	to obtain the result. The proof is completed.
	
\end{proof}

\begin{proof}[\textit{\textbf{Proof of Theorem \ref{theorem_main2}}}]
	As the logistic loss is $ 1 $-Lipschitz, the log-likelihood satisfies that 
	$
	\left\vert\log f(x)-\log f(y) \right\vert 
	\leq 
	|x-y|
	$.
	Thus, we can deduce that
	\begin{align}
	\mathbb{E}\left[\log
	\left(
	\frac{p_M}{ p_{M^*} }
	\right)^2 \right]
	& \leq 
	\frac{1}{d_1  d_2} \sum_{i\in[d_1]}
	\sum_{j\in[d_2]} \Pi_{ij} 
	\left(\log f(M^*_{ij}) -\log f(M_{ij})\right)^2   \nonumber
	\\
	& \leq 
	\frac{1}{d_1  d_2} \sum_{i\in[d_1]}
	\sum_{j\in[d_2]} 
	\left(\log f(M^*_{ij}) -\log f(M_{ij})\right)^2   \nonumber
	\\
	& \leq 
	\frac{1}{d_1  d_2} \sum_{i\in[d_1]}
	\sum_{j\in[d_2]} 
	\left( M^*_{ij}- M_{ij}\right)^2    \nonumber
	\\
	& \leq 
	\frac{1}{d_1  d_2}
	\| M^* -M \|^2_F
	\label{eq_bound_KL2222}
	,	
	\end{align}
	where $ \Pi_{ij}\leq 1 $ is the probability to observe the $ (i,j) $-th entry.
	From \eqref{eq_klbound_1}, we have that
	\begin{align*}
	\mathcal{K} (P_{M^*}, P_M) 
	\leq
	\frac{\| M^* -M \|_F }{ \sqrt{ d_1 d_2} }
	\leq 	\sqrt{ \delta 2^{3} d_1 d_2  B}
	,
	\end{align*}
	and
	\begin{align*}
	\mathbb{E}\left[\log
	\left(
	\frac{p_M}{ p_{M^*} }
	\right)^2 \right]
	\leq
	\frac{\| M^* -M \|^2_F}{d_1  d_2}
	\leq 
	\delta 2^{3} d_1 d_2  B
	.
	\end{align*}	
	For any $(U,V)$ in the support of $\rho_n$, given in \eqref{eq_specific_rhon},  we observe that for $\delta=\frac{B}{8(n d_1 d_2)^2}$, where $\delta$ satisfies $0 < \delta < B$, and from equation \eqref{eq_bound_Frobenius} we can deduce that
	\begin{align*}
	\int \mathcal{K} (P_{M^*}, P_M) \rho_n(dM)
	\leq
	\frac{B}{n\sqrt{ d_1 d_2}},
	\end{align*}
	and
	\begin{align*}
	\int \mathbb{E}\left[\log^2 \left( \frac{p_M}{ p_{M^*} }\right) \right]  \rho_n(dM)
	\leq
	\frac{B^2}{n^2 d_1 d_2}
	.
	\end{align*}	
	Now, from Lemma \ref{lm_boundKL}, we have that
	\begin{equation*}
	\frac{1}{n}\mathcal{K}(\rho_n,\pi) 
	\leq 
	\frac{
		2(1+2a) r(d_1 + d_2) \left[ \log(nd_1 d_2)
		+
		C_a
		\right] }{n}
	.
	\end{equation*}	
	We now can apply Corollary 2.5 and Theorem 2.4 in \cite{alquier2020concentration} with $ \rho_n $ in \eqref{eq_specific_rhon} and
	\begin{equation*}
	\varepsilon_n = C_{a,B}\frac{ r(d_1+d_2)\log(nd_1d_2)}{n} 	
	\end{equation*}	
	to obtain the result. The proof is completed.
	
\end{proof}

\begin{proof}[\textit{\textbf{Proof of Corollary \ref{cor_concentration_Hellinger}}}]
	From \cite{van2014renyi}, we have that 
	$$ 
	H^2(P,Q)
	\leq 
	D_{1/2}(P,Q)  
	\leq 
	D_{\alpha}(P,Q) 
	,
	$$
	for $ \alpha \in [0.5,1) $. In addition, we also have that 
	$$ 
	D_{1/2}(P,Q)  
	\leq 
	\frac{(1-\alpha)1/2}{\alpha (1-1/2)} D_{\alpha}(P,Q) 
	=
	\frac{(1-\alpha)}{\alpha} D_{\alpha}(P,Q) 
	,
	$$ 
	for $ \alpha \in (0, 0.5)  $.
	
	Thus, using definition of $ c_\alpha  $ and Theorem \ref{theorem_main2}, we obtain the results.
	
\end{proof}

\begin{proof}[\textit{\textbf{Proof of Theorem \ref{theorem_estimation}}}]
	From \eqref{eq_Hellinger_mainresults}, we have that 
	\begin{align*}
	\mathbb{P}\left[
	\int H^2(P_M ,P_{M^*} ) \pi_{n,\alpha}({\rm d} M )  
	\leq 
	c_\alpha 
	\varepsilon_n \right]
	\geq 
	1-\frac{2}{n\varepsilon_n}
	,
	\end{align*}	
	from Lemma \ref{lm_Frobento_heling}, one has that
	\begin{align*}
	\mathbb{P}\left[
	\int 
	\frac{C_1 C_\kappa \| M - M^* \|^2_F
	}{d_1 d_2}
	\pi_{n,\alpha}({\rm d} M )  
	\leq 
	c_\alpha 
	\varepsilon_n \right]
	\geq 
	1-\frac{2}{n\varepsilon_n}
	,
	\end{align*}
	thus, we obtain \eqref{eq_estimation_concentration}. 
	To obtain \eqref{eq_bound_meanestimator}, one can apple Jensen's inequality for a convex function, that
	\begin{equation*}
	\| \hat{M} - M^* \|^2_F
	=
	\bigg\| \int M \pi_{n,\alpha}({\rm d}M)  - M^* \bigg\|^2_F
	\leq
	\int 
	\| M- M^* \|^2_F
	\pi_{n,\alpha}({\rm d}M) 
	. 
	\end{equation*}
	This completes the proof.
	
\end{proof}

\subsection{Proofs for Section \ref{sc_main2}}
\label{sc_proofs2}
\begin{proof}[\textit{\textbf{Proof of Theorem \ref{theorem_main_prior2}}}]
	From \eqref{eq_klbound_1}, we have that
	\begin{align*}
	\mathcal{K} (P_{M^*}, P_M) 
	\leq
	\frac{\| M^* -M \|_F }{ \sqrt{ d_1 d_2} }
	,
	\end{align*}
	When integrating with respect to $\rho_n:= \rho_{0} $ given in \eqref{eq_priorspecific_anark}, we have that
	\begin{align}
	\int \mathcal{K} (P_{M^*}, P_M) 
	\rho_n(dM) \nonumber
	& \leq
	\int
	\frac{\| M^* -M \|_F }{ \sqrt{ d_1 d_2} }
	\rho_0(dM)  \nonumber
	\\
	& =
	\int
	\frac{\| M^* -M \|_F }{ \sqrt{ d_1 d_2} }
	\pi_{st} (M - M^*)dM   \nonumber
	\\
	& =
	\frac{1}{ \sqrt{ d_1 d_2} }
	\int  \| M \|_F \pi_{st} (M)dM  \nonumber
	\\
	& \leq
	\frac{1}{ \sqrt{ d_1 d_2} }
	\left(\int  \| M \|_F^2 \pi_{st}(M)dM \right)^{1/2}  \nonumber
	\\
	& \leq
	\frac{1}{ \sqrt{ d_1 d_2} } \sqrt{d_1 d_2 \tau^2}
	=
	\tau
	\label{eq_boundKLarnak}
	,
	\end{align}
	where we have used Holder's inequality and Lemma \ref{lemma:arnak:1} to obtain the result.
	Now, from Lemma \ref{lemma:arnak:2}, we have that
	\begin{equation*}
	\frac{1}{n}\mathcal{K}(\rho_n,\pi_{st}) 
	\leq 
	\frac{
		2 r (d_1 + d_2 +2) \log \left( 1+ \frac{\| M^* \|_F}{\tau \sqrt{2r }} \right) 
	}{n}
	.
	\end{equation*}	
	Taking $ \tau = 1/n $, we obtain that
	\begin{align*}
	&	\int \mathcal{K} (P_{M^*}, P_M) 
	\rho_n(dM)
	\leq 
	\frac{1}{n}
	,
	\end{align*}
	\begin{align*}
	\frac{1}{n}\mathcal{K}(\rho_n,\pi_{st}) 
	\leq 
	\frac{
		2 r (d_1 + d_2 +2) \log \left( 1+ \frac{ n \| M^* \|_F}{ \sqrt{2r }} \right) 
	}{n}
	.
	\end{align*}	
	We now can apply Theorem 2.6 in \cite{alquier2020concentration} with 
	$$ 
	\varepsilon_n = \frac{
		2 r (d_1 + d_2 +2) \log \left( 1+ \frac{ n \| M^* \|_F}{\sqrt{2r }} \right) 
	}{n} 
	$$ 
	to obtain the result. The proof is completed.
	
\end{proof}

\begin{proof}[\textit{\textbf{Proof of Theorem \ref{theorem_main_prior2_prob}}}]
	From \eqref{eq_klbound_1}, we have that
	\begin{align*}
	\mathcal{K} (P_{M^*}, P_M) 
	\leq
	\frac{\| M^* -M \|_F }{ \sqrt{ d_1 d_2} }
	,
	\end{align*}
	When integrating with respect to $\rho_n:= \rho_{0} $ given in \eqref{eq_priorspecific_anark}, and from \eqref{eq_boundKLarnak}, we have that
	\begin{align*}
	\int \mathcal{K} (P_{M^*}, P_M) 
	\rho_n(dM)
	\leq
	\tau
	.
	\end{align*}
	Now, from Lemma \ref{lemma:arnak:2}, we have that
	\begin{equation*}
	\frac{1}{n}\mathcal{K}(\rho_n,\pi_{st}) 
	\leq 
	\frac{
		2 r (d_1 + d_2 +2) \log \left( 1+ \frac{\| M^* \|_F}{\tau \sqrt{2r }} \right) 
	}{n}
	.
	\end{equation*}	
	Moreover, from \eqref{eq_bound_KL2222}, one has that
	\begin{align*}
	\mathbb{E}\left[\log
	\left(
	\frac{p_M}{ p_{M^*} }
	\right)^2 \right]
	\leq
	\frac{\| M -M^*  \|^2_F}{d_1  d_2}
	,
	\end{align*}
	and when integrating with respect to $\rho_n:= \rho_{0} $ given in \eqref{eq_priorspecific_anark}, it leads to
	\begin{align*}
	\int \mathbb{E}\left[\log
	\left(
	\frac{p_M}{ p_{M^*} }
	\right)^2 \right] 	\rho_n(dM)
	& \leq
	\int \frac{\| M -M^*  \|^2_F}{d_1  d_2} 	\rho_n(dM)
	\\
	& =
	\int \frac{\| M-M^*  \|^2_F}{d_1  d_2} 
	\pi_{st} (M - M^*)dM  
	\\
	& =	
	\frac{1}{d_1  d_2} \int  \| M \|_F^2 \pi_{st}(M)dM
	\\
	& \leq	
	\frac{d_1  d_2 \tau^2}{d_1  d_2} 
	=
	\tau^2
	,
	\end{align*}
	where we have used a change of variable and Lemma \ref{lemma:arnak:1} to obtain the result.	
	
	Now, by taking $ \tau = 1/n $, we obtain that
	\begin{align*}
	\int \mathcal{K} (P_{M^*}, P_M) 
	\rho_n(dM)
	&	\leq 
	\frac{1}{n},
	\\
	\int \mathbb{E}\left[\log
	\left(
	\frac{p_M}{ p_{M^*} }
	\right)^2 \right] 	\rho_n(dM)	
	&	\leq 
	\frac{1}{n^2},
	\\
	\frac{1}{n}\mathcal{K}(\rho_n,\pi_{st}) 
	&	\leq 
	\frac{
		2 r (d_1 + d_2 +2) \log \left( 1+ \frac{ n \| M^* \|_F}{  \sqrt{2r }} \right) 
	}{n}
	.
	\end{align*}	
	We now can apply Theorem 2.4 and Corollary 2.5 in \cite{alquier2020concentration} with 
	$$ 
	\varepsilon_n = \frac{
		2 r (d_1 + d_2 +2) \log \left( 1+ \frac{ n \| M^* \|_F}{  \sqrt{2r }} \right) 
	}{n} 
	$$ 
	to obtain the result. The proof is completed.
	
\end{proof}

\begin{proof}[\textit{\textbf{Proof of Theorem \ref{theorem_estimation2}}}]
	From Theorem  \ref{theorem_main_prior2_prob}, using a bound for Hellinger distance as in Corollary \ref{cor_concentration_Hellinger}, we have that 
	\begin{align*}
	\mathbb{P}\left[
	\int H^2(P_M ,P_{M^*} ) \pi_{n,\alpha}({\rm d} M )  
	\leq 
	c_\alpha 
	\varepsilon_n \right]
	\geq 
	1-\frac{2}{n\varepsilon_n}
	,
	\end{align*}	
	from Lemma \ref{lm_Frobento_heling}, it yields that
	\begin{align*}
	\mathbb{P}\left[
	\int 
	\frac{C_1 C_\kappa \| M - M^* \|^2_F
	}{d_1 d_2}
	\pi_{n,\alpha}({\rm d} M )  
	\leq 
	c_\alpha 
	\varepsilon_n \right]
	\geq 
	1-\frac{2}{n\varepsilon_n}
	,
	\end{align*}
	thus, we obtain \eqref{eq_estimation_concentration2}. 
	To obtain \eqref{eq_bound_meanestimator2}, one can apple Jensen's inequality for a convex function, that
	\begin{equation*}
	\| \hat{M} - M^* \|^2_F
	=
	\bigg\| \int M \pi_{n,\alpha}({\rm d}M)  - M^* \bigg\|^2_F
	\leq
	\int 
	\| M- M^* \|^2_F
	\pi_{n,\alpha}({\rm d}M) 
	,
	\end{equation*}
	and combine with result in \eqref{eq_estimation_concentration2}.
	This completes the proof.
	
\end{proof}

\subsection{Lemma}

\begin{definition}
Fix $B>0$, $r\geq 1$. For any pair $(\bar{U},\bar{V})\in\mathcal{M}(r,B) $, we define for $\delta\in(0,B)$ that will be chosen later,
\begin{equation}
\label{eq_specific_rhon}
\rho_n({\rm d}U,{\rm d}V,{\rm d}\gamma) 
\propto 
\mathbf{1}_{(\|U-\bar{U}\|_{\infty} \leq \delta,\|U-\bar{U}\|_{\infty} \leq \delta)} \pi({\rm d}U,{\rm d}V,{\rm d}\gamma).
\end{equation}
\end{definition}

\begin{lemma}
	\label{lm_boundKL}
	Put $ C_a:= \log(8\sqrt{\pi}\Gamma(a)2^{10a+1})+3 $
	and with $\delta=B/[8(nd_1 d_2)^2]$ that satisfies $0<\delta<B$, we have for $ \rho_n $ in \eqref{eq_specific_rhon} that
	\begin{equation*}
	\mathcal{K}(\rho_n,\pi) \leq 2(1+2a) r(d_1 + d_2) \left[ \log(nd_1 d_2)
	+
	C_a
	\right].
	\end{equation*}	
\end{lemma}
\begin{proof}[\textit{\textbf{Proof of Lemma \ref{lm_boundKL}}}]
	This result can found, for example, in the proof of Theorem 4.1 in \cite{alquier2020concentration}.
	
\end{proof}

\begin{lemma}
	\label{lm_Frobento_heling}
	For any matrix $ A \in \mathbb{R}^{d_1 \times d_2} $ and $ B \in \mathbb{R}^{d_1 \times d_2} $ satisfying that $ \|A\|_\infty \leq \kappa $ and $ \|B\|_\infty \leq \kappa $, under Assumption \ref{assum_boundedness} and Assumption \ref{assum_observation_prob}, one has that
	\begin{equation*}
	\frac{ \| A - B \|^2_F
	}{d_1 d_2}
	\leq
	\frac{ H^2(P_{A},P_{B}) }{C_1 C_\kappa}
	.
	\end{equation*}
\end{lemma}
\begin{proof}[\textit{\textbf{Proof of Lemma \ref{lm_Frobento_heling}}}]
	This is Lemma A.2 in \cite{Davenport14}. With $ d^2_H (p,q) := (\sqrt{p} -\sqrt{q} )^2 + (\sqrt{1-p} -\sqrt{1-q} )^2 $ for two number $ p,q \in [0,1] $, it is noting under Assumption \ref{assum_observation_prob} that 
	\begin{align*}
	H^2(P_{A},P_{B})
	&	=
	\frac{1}{d_1 d_2} \sum_{i\in[d_1]}
	\sum_{j\in[d_2]} \Pi_{ij} 
	d^2_H (f(A_{ij}),f(B_{ij}))
	\\
	&	\geq
	\frac{C_1}{d_1 d_2} \sum_{i\in[d_1]}
	\sum_{j\in[d_2]} d^2_H (f(A_{ij}),f(B_{ij}))
		\\
	&	\geq
	C_1
	H^2(f(A),f(B))
	.
	\end{align*}
	where $ \Pi_{ij} $ is the probability to observe the $ (i,j) $-th entry. Now from Lemma A.2 in \cite{Davenport14}, under Assumption \ref{assum_boundedness}, one has that
\begin{align*}
	H^2(f(A),f(B))
	\geq
	C_\kappa	\frac{ \| A - B \|^2_F
	}{d_1 d_2}.
\end{align*}	
 The argument is also similar to Lemma 9 and Lemma 11 in \cite{klopp2015adaptive}. This completes the proof.
	
\end{proof}
Finally, we will use quite often the following distribution that will be defined as translations of the prior $\pi_{st} $ in \eqref{prior_scaled_Student}. We introduce the following notation.
\begin{definition}
	\label{definition:posterior:transla}
	Let's define 
	\begin{equation}
	\label{eq_priorspecific_anark}
	\rho_{0}(M) = \pi_{st} (M - M^*)
	.	
	\end{equation}
\end{definition}
\noindent The following technical lemmas will be useful in the proofs.

\begin{lemma}[Lemma 1 in \cite{dalalyan2020exponential}]
	\label{lemma:arnak:1}
	We have
	$$
	\int \| M \|_{F}^2 \pi_{st} ( M ){\rm d} M  
	\leq 
	d_1 d_2 \tau^2
	. 
	$$
\end{lemma}

\begin{lemma}[Lemma 2 in \cite{dalalyan2020exponential}]
	\label{lemma:arnak:2}
	We have
	$$
	KL( \rho_{0 } , \pi_{st}) 
	\leq 
	2 r (d_1 + d_2 +2) \log \left( 1+ \frac{\| M^* \|_F}{\tau \sqrt{2r }} \right) 
	,
	$$
	with the convention $0\log(1+0/0)=0$.
\end{lemma}

\bmhead{Acknowledgements}
The author was supported by the Norwegian Research Council, grant number 309960, through the Centre for Geophysical Forecasting at NTNU.

\subsubsection*{Conflict of interest/Competing interests:}
 The author declares no potential conflict of interests.






\begin{thebibliography}{58}
	\ifx \bisbn   \undefined \def \bisbn  #1{ISBN #1}\fi
	\ifx \binits  \undefined \def \binits#1{#1}\fi
	\ifx \bauthor  \undefined \def \bauthor#1{#1}\fi
	\ifx \batitle  \undefined \def \batitle#1{#1}\fi
	\ifx \bjtitle  \undefined \def \bjtitle#1{#1}\fi
	\ifx \bvolume  \undefined \def \bvolume#1{\textbf{#1}}\fi
	\ifx \byear  \undefined \def \byear#1{#1}\fi
	\ifx \bissue  \undefined \def \bissue#1{#1}\fi
	\ifx \bfpage  \undefined \def \bfpage#1{#1}\fi
	\ifx \blpage  \undefined \def \blpage #1{#1}\fi
	\ifx \burl  \undefined \def \burl#1{\textsf{#1}}\fi
	\ifx \doiurl  \undefined \def \doiurl#1{\url{https://doi.org/#1}}\fi
	\ifx \betal  \undefined \def \betal{\textit{et al.}}\fi
	\ifx \binstitute  \undefined \def \binstitute#1{#1}\fi
	\ifx \binstitutionaled  \undefined \def \binstitutionaled#1{#1}\fi
	\ifx \bctitle  \undefined \def \bctitle#1{#1}\fi
	\ifx \beditor  \undefined \def \beditor#1{#1}\fi
	\ifx \bpublisher  \undefined \def \bpublisher#1{#1}\fi
	\ifx \bbtitle  \undefined \def \bbtitle#1{#1}\fi
	\ifx \bedition  \undefined \def \bedition#1{#1}\fi
	\ifx \bseriesno  \undefined \def \bseriesno#1{#1}\fi
	\ifx \blocation  \undefined \def \blocation#1{#1}\fi
	\ifx \bsertitle  \undefined \def \bsertitle#1{#1}\fi
	\ifx \bsnm \undefined \def \bsnm#1{#1}\fi
	\ifx \bsuffix \undefined \def \bsuffix#1{#1}\fi
	\ifx \bparticle \undefined \def \bparticle#1{#1}\fi
	\ifx \barticle \undefined \def \barticle#1{#1}\fi
	\bibcommenthead
	\ifx \bconfdate \undefined \def \bconfdate #1{#1}\fi
	\ifx \botherref \undefined \def \botherref #1{#1}\fi
	\ifx \url \undefined \def \url#1{\textsf{#1}}\fi
	\ifx \bchapter \undefined \def \bchapter#1{#1}\fi
	\ifx \bbook \undefined \def \bbook#1{#1}\fi
	\ifx \bcomment \undefined \def \bcomment#1{#1}\fi
	\ifx \oauthor \undefined \def \oauthor#1{#1}\fi
	\ifx \citeauthoryear \undefined \def \citeauthoryear#1{#1}\fi
	\ifx \endbibitem  \undefined \def \endbibitem {}\fi
	\ifx \bconflocation  \undefined \def \bconflocation#1{#1}\fi
	\ifx \arxivurl  \undefined \def \arxivurl#1{\textsf{#1}}\fi
	\csname PreBibitemsHook\endcsname
	
	%%% 1
	\bibitem[\protect\citeauthoryear{Alquier et~al.}{2014}]{alquier2014bayesian}
	\begin{botherref}
		\oauthor{\bsnm{Alquier}, \binits{P.}},
		\oauthor{\bsnm{Cottet}, \binits{V.}},
		\oauthor{\bsnm{Chopin}, \binits{N.}},
		\oauthor{\bsnm{Rousseau}, \binits{J.}}:
		{B}ayesian matrix completion: prior specification and consistency.
		arXiv preprint arXiv:1406.1440
		(2014)
	\end{botherref}
	\endbibitem
	
	%%% 2
	\bibitem[\protect\citeauthoryear{Alquier et~al.}{2019}]{alquier2019estimation}
	\begin{barticle}
		\bauthor{\bsnm{Alquier}, \binits{P.}},
		\bauthor{\bsnm{Cottet}, \binits{V.}},
		\bauthor{\bsnm{Lecu{\'e}}, \binits{G.}}:
		\batitle{Estimation bounds and sharp oracle inequalities of regularized
			procedures with lipschitz loss functions}.
		\bjtitle{The Annals of Statistics}
		\bvolume{47}(\bissue{4}),
		\bfpage{2117}--\blpage{2144}
		(\byear{2019})
	\end{barticle}
	\endbibitem
	
	%%% 3
	\bibitem[\protect\citeauthoryear{Alaya and Klopp}{2019}]{alaya2019collective}
	\begin{barticle}
		\bauthor{\bsnm{Alaya}, \binits{M.Z.}},
		\bauthor{\bsnm{Klopp}, \binits{O.}}:
		\batitle{Collective matrix completion.}
		\bjtitle{J. Mach. Learn. Res.}
		\bvolume{20},
		\bfpage{148}--\blpage{1}
		(\byear{2019})
	\end{barticle}
	\endbibitem
	
	%%% 4
	\bibitem[\protect\citeauthoryear{Alquier and
		Ridgway}{2020}]{alquier2020concentration}
	\begin{barticle}
		\bauthor{\bsnm{Alquier}, \binits{P.}},
		\bauthor{\bsnm{Ridgway}, \binits{J.}}:
		\batitle{Concentration of tempered posteriors and of their variational
			approximations}.
		\bjtitle{The Annals of Statistics}
		\bvolume{48}(\bissue{3}),
		\bfpage{1475}--\blpage{1497}
		(\byear{2020})
	\end{barticle}
	\endbibitem
	
	%%% 5
	\bibitem[\protect\citeauthoryear{Bissiri et~al.}{2016}]{bissiri2013general}
	\begin{barticle}
		\bauthor{\bsnm{Bissiri}, \binits{P.G.}},
		\bauthor{\bsnm{Holmes}, \binits{C.C.}},
		\bauthor{\bsnm{Walker}, \binits{S.G.}}:
		\batitle{A general framework for updating belief distributions}.
		\bjtitle{Journal of the Royal Statistical Society: Series B (Statistical
			Methodology)}
		\bvolume{78}(\bissue{5}),
		\bfpage{1103}--\blpage{1130}
		(\byear{2016})
	\end{barticle}
	\endbibitem
	
	%%% 6
	\bibitem[\protect\citeauthoryear{Bennett and
		Lanning}{2007}]{bennett2007netflix}
	\begin{bchapter}
		\bauthor{\bsnm{Bennett}, \binits{J.}},
		\bauthor{\bsnm{Lanning}, \binits{S.}}:
		\bctitle{The netflix prize}.
		In: \bbtitle{Proceedings of KDD Cup and Workshop},
		vol. \bseriesno{2007},
		p. \bfpage{35}
		(\byear{2007})
	\end{bchapter}
	\endbibitem
	
	%%% 7
	\bibitem[\protect\citeauthoryear{Babacan et~al.}{2012}]{babacan2012sparse}
	\begin{barticle}
		\bauthor{\bsnm{Babacan}, \binits{S.D.}},
		\bauthor{\bsnm{Luessi}, \binits{M.}},
		\bauthor{\bsnm{Molina}, \binits{R.}},
		\bauthor{\bsnm{Katsaggelos}, \binits{A.K.}}:
		\batitle{Sparse {B}ayesian methods for low-rank matrix estimation}.
		\bjtitle{IEEE Transactions on Signal Processing}
		\bvolume{60}(\bissue{8}),
		\bfpage{3964}--\blpage{3977}
		(\byear{2012})
	\end{barticle}
	\endbibitem
	
	%%% 8
	\bibitem[\protect\citeauthoryear{Bobadilla
		et~al.}{2013}]{bobadilla2013recommender}
	\begin{barticle}
		\bauthor{\bsnm{Bobadilla}, \binits{J.}},
		\bauthor{\bsnm{Ortega}, \binits{F.}},
		\bauthor{\bsnm{Hernando}, \binits{A.}},
		\bauthor{\bsnm{Guti{\'e}rrez}, \binits{A.}}:
		\batitle{Recommender systems survey}.
		\bjtitle{Knowledge-based systems}
		\bvolume{46},
		\bfpage{109}--\blpage{132}
		(\byear{2013})
	\end{barticle}
	\endbibitem
	
	%%% 9
	\bibitem[\protect\citeauthoryear{Bhattacharya
		et~al.}{2019}]{bhattacharya2016bayesian}
	\begin{barticle}
		\bauthor{\bsnm{Bhattacharya}, \binits{A.}},
		\bauthor{\bsnm{Pati}, \binits{D.}},
		\bauthor{\bsnm{Yang}, \binits{Y.}}:
		\batitle{Bayesian fractional posteriors}.
		\bjtitle{Annals of Statistics}
		\bvolume{47}(\bissue{1}),
		\bfpage{39}--\blpage{66}
		(\byear{2019})
	\end{barticle}
	\endbibitem
	
	%%% 10
	\bibitem[\protect\citeauthoryear{Cottet and Alquier}{2018}]{cottet20181bit}
	\begin{barticle}
		\bauthor{\bsnm{Cottet}, \binits{V.}},
		\bauthor{\bsnm{Alquier}, \binits{P.}}:
		\batitle{1-bit matrix completion: Pac-bayesian analysis of a variational
			approximation}.
		\bjtitle{Machine Learning}
		\bvolume{107}(\bissue{3}),
		\bfpage{579}--\blpage{603}
		(\byear{2018})
	\end{barticle}
	\endbibitem
	
	%%% 11
	\bibitem[\protect\citeauthoryear{Chen et~al.}{2019}]{chen2019inference}
	\begin{barticle}
		\bauthor{\bsnm{Chen}, \binits{Y.}},
		\bauthor{\bsnm{Fan}, \binits{J.}},
		\bauthor{\bsnm{Ma}, \binits{C.}},
		\bauthor{\bsnm{Yan}, \binits{Y.}}:
		\batitle{Inference and uncertainty quantification for noisy matrix completion}.
		\bjtitle{Proceedings of the National Academy of Sciences}
		\bvolume{116}(\bissue{46}),
		\bfpage{22931}--\blpage{22937}
		(\byear{2019})
	\end{barticle}
	\endbibitem
	
	%%% 12
	\bibitem[\protect\citeauthoryear{Chatterjee}{2015}]{chatterjee2015matrix}
	\begin{barticle}
		\bauthor{\bsnm{Chatterjee}, \binits{S.}}:
		\batitle{Matrix estimation by universal singular value thresholding}.
		\bjtitle{The Annals of Statistics}
		\bvolume{43}(\bissue{1}),
		\bfpage{177}--\blpage{214}
		(\byear{2015})
	\end{barticle}
	\endbibitem
	
	%%% 13
	\bibitem[\protect\citeauthoryear{Candes and Plan}{2010}]{candes2010matrix}
	\begin{barticle}
		\bauthor{\bsnm{Candes}, \binits{E.J.}},
		\bauthor{\bsnm{Plan}, \binits{Y.}}:
		\batitle{Matrix completion with noise}.
		\bjtitle{Proceedings of the IEEE}
		\bvolume{98}(\bissue{6}),
		\bfpage{925}--\blpage{936}
		(\byear{2010})
	\end{barticle}
	\endbibitem
	
	%%% 14
	\bibitem[\protect\citeauthoryear{Cand{\`e}s and Recht}{2009}]{CandesRecht}
	\begin{barticle}
		\bauthor{\bsnm{Cand{\`e}s}, \binits{E.J.}},
		\bauthor{\bsnm{Recht}, \binits{B.}}:
		\batitle{Exact matrix completion via convex optimization}.
		\bjtitle{Found. Comput. Math.}
		\bvolume{9}(\bissue{6}),
		\bfpage{717}--\blpage{772}
		(\byear{2009})
		\doiurl{10.1007/s10208-009-9045-5}
	\end{barticle}
	\endbibitem
	
	%%% 15
	\bibitem[\protect\citeauthoryear{Cand{\`e}s and Tao}{2010}]{CandesT10}
	\begin{barticle}
		\bauthor{\bsnm{Cand{\`e}s}, \binits{E.J.}},
		\bauthor{\bsnm{Tao}, \binits{T.}}:
		\batitle{The power of convex relaxation: near-optimal matrix completion}.
		\bjtitle{IEEE Trans. Inform. Theory}
		\bvolume{56}(\bissue{5}),
		\bfpage{2053}--\blpage{2080}
		(\byear{2010})
		\doiurl{10.1109/TIT.2010.2044061}
	\end{barticle}
	\endbibitem
	
	%%% 16
	\bibitem[\protect\citeauthoryear{Cai and Zhou}{2013}]{cai2013}
	\begin{barticle}
		\bauthor{\bsnm{Cai}, \binits{T.}},
		\bauthor{\bsnm{Zhou}, \binits{W.-X.}}:
		\batitle{A max-norm constrained minimization approach to 1-bit matrix
			completion.}
		\bjtitle{J. Mach. Learn. Res.}
		\bvolume{14}(\bissue{1}),
		\bfpage{3619}--\blpage{3647}
		(\byear{2013})
	\end{barticle}
	\endbibitem
	
	%%% 17
	\bibitem[\protect\citeauthoryear{Chi et~al.}{2013}]{chi2013genotype}
	\begin{barticle}
		\bauthor{\bsnm{Chi}, \binits{E.C.}},
		\bauthor{\bsnm{Zhou}, \binits{H.}},
		\bauthor{\bsnm{Chen}, \binits{G.K.}},
		\bauthor{\bsnm{Del~Vecchyo}, \binits{D.O.}},
		\bauthor{\bsnm{Lange}, \binits{K.}}:
		\batitle{Genotype imputation via matrix completion}.
		\bjtitle{Genome research}
		\bvolume{23}(\bissue{3}),
		\bfpage{509}--\blpage{518}
		(\byear{2013})
	\end{barticle}
	\endbibitem
	
	%%% 18
	\bibitem[\protect\citeauthoryear{Dalalyan}{2017}]{dalalyan2017theoretical}
	\begin{barticle}
		\bauthor{\bsnm{Dalalyan}, \binits{A.S.}}:
		\batitle{Theoretical guarantees for approximate sampling from smooth and
			log-concave densities}.
		\bjtitle{Journal of the Royal Statistical Society: Series B Statistical
			Methodology}
		\bvolume{79}(\bissue{3}),
		\bfpage{651}--\blpage{676}
		(\byear{2017})
	\end{barticle}
	\endbibitem
	
	%%% 19
	\bibitem[\protect\citeauthoryear{Dalalyan}{2020}]{dalalyan2020exponential}
	\begin{barticle}
		\bauthor{\bsnm{Dalalyan}, \binits{A.S.}}:
		\batitle{Exponential weights in multivariate regression and a low-rankness
			favoring prior}.
		\bjtitle{Annales de l'Institut Henri Poincar{\'e}, Probabilit{\'e}s et
			Statistiques}
		\bvolume{56}(\bissue{2}),
		\bfpage{1465}--\blpage{1483}
		(\byear{2020})
	\end{barticle}
	\endbibitem
	
	%%% 20
	\bibitem[\protect\citeauthoryear{Dalalyan and
		Karagulyan}{2019}]{dalalyan2019user}
	\begin{barticle}
		\bauthor{\bsnm{Dalalyan}, \binits{A.S.}},
		\bauthor{\bsnm{Karagulyan}, \binits{A.}}:
		\batitle{User-friendly guarantees for the langevin monte carlo with inaccurate
			gradient}.
		\bjtitle{Stochastic Processes and their Applications}
		\bvolume{129}(\bissue{12}),
		\bfpage{5278}--\blpage{5311}
		(\byear{2019})
	\end{barticle}
	\endbibitem
	
	%%% 21
	\bibitem[\protect\citeauthoryear{Durmus and
		Moulines}{2017}]{durmus2017nonasymptotic}
	\begin{barticle}
		\bauthor{\bsnm{Durmus}, \binits{A.}},
		\bauthor{\bsnm{Moulines}, \binits{E.}}:
		\batitle{Nonasymptotic convergence analysis for the unadjusted langevin
			algorithm}.
		\bjtitle{The Annals of Applied Probability}
		\bvolume{27}(\bissue{3}),
		\bfpage{1551}--\blpage{1587}
		(\byear{2017})
	\end{barticle}
	\endbibitem
	
	%%% 22
	\bibitem[\protect\citeauthoryear{Durmus and Moulines}{2019}]{durmus2019high}
	\begin{barticle}
		\bauthor{\bsnm{Durmus}, \binits{A.}},
		\bauthor{\bsnm{Moulines}, \binits{E.}}:
		\batitle{High-dimensional {B}ayesian inference via the unadjusted langevin
			algorithm}.
		\bjtitle{Bernoulli}
		\bvolume{25}(\bissue{4A}),
		\bfpage{2854}--\blpage{2882}
		(\byear{2019})
	\end{barticle}
	\endbibitem
	
	%%% 23
	\bibitem[\protect\citeauthoryear{Davenport et~al.}{2014}]{Davenport14}
	\begin{barticle}
		\bauthor{\bsnm{Davenport}, \binits{M.A.}},
		\bauthor{\bsnm{Plan}, \binits{Y.}},
		\bauthor{\bsnm{Van Den~Berg}, \binits{E.}},
		\bauthor{\bsnm{Wootters}, \binits{M.}}:
		\batitle{1-bit matrix completion}.
		\bjtitle{Information and Inference: A Journal of the IMA}
		\bvolume{3}(\bissue{3}),
		\bfpage{189}--\blpage{223}
		(\byear{2014})
	\end{barticle}
	\endbibitem
	
	%%% 24
	\bibitem[\protect\citeauthoryear{Dalalyan and
		Tsybakov}{2012a}]{dalalyan2012mirror}
	\begin{barticle}
		\bauthor{\bsnm{Dalalyan}, \binits{A.S.}},
		\bauthor{\bsnm{Tsybakov}, \binits{A.}}:
		\batitle{Mirror averaging with sparsity priors}.
		\bjtitle{Bernoulli}
		\bvolume{18}(\bissue{3}),
		\bfpage{914}--\blpage{944}
		(\byear{2012})
	\end{barticle}
	\endbibitem
	
	%%% 25
	\bibitem[\protect\citeauthoryear{Dalalyan and
		Tsybakov}{2012b}]{dalalyan2012sparse}
	\begin{barticle}
		\bauthor{\bsnm{Dalalyan}, \binits{A.S.}},
		\bauthor{\bsnm{Tsybakov}, \binits{A.B.}}:
		\batitle{Sparse regression learning by aggregation and langevin monte-carlo}.
		\bjtitle{Journal of Computer and System Sciences}
		\bvolume{78}(\bissue{5}),
		\bfpage{1423}--\blpage{1443}
		(\byear{2012})
	\end{barticle}
	\endbibitem
	
	%%% 26
	\bibitem[\protect\citeauthoryear{Gross}{2011}]{gross2011recovering}
	\begin{barticle}
		\bauthor{\bsnm{Gross}, \binits{D.}}:
		\batitle{Recovering low-rank matrices from few coefficients in any basis}.
		\bjtitle{IEEE Transactions on Information Theory}
		\bvolume{57}(\bissue{3}),
		\bfpage{1548}--\blpage{1566}
		(\byear{2011})
	\end{barticle}
	\endbibitem
	
	%%% 27
	\bibitem[\protect\citeauthoryear{Gr{\"u}nwald and
		Van~Ommen}{2017}]{grunwald2017inconsistency}
	\begin{barticle}
		\bauthor{\bsnm{Gr{\"u}nwald}, \binits{P.}},
		\bauthor{\bsnm{Van~Ommen}, \binits{T.}}:
		\batitle{Inconsistency of {B}ayesian inference for misspecified linear models,
			and a proposal for repairing it}.
		\bjtitle{{B}ayesian Analysis}
		\bvolume{12}(\bissue{4}),
		\bfpage{1069}--\blpage{1103}
		(\byear{2017})
	\end{barticle}
	\endbibitem
	
	%%% 28
	\bibitem[\protect\citeauthoryear{Hong and Martin}{2020}]{hong2020model}
	\begin{barticle}
		\bauthor{\bsnm{Hong}, \binits{L.}},
		\bauthor{\bsnm{Martin}, \binits{R.}}:
		\batitle{{Model misspecification, Bayesian versus credibility estimation, and
				Gibbs posteriors}}.
		\bjtitle{Scandinavian Actuarial Journal}
		\bvolume{2020}(\bissue{7}),
		\bfpage{634}--\blpage{649}
		(\byear{2020})
	\end{barticle}
	\endbibitem
	
	%%% 29
	\bibitem[\protect\citeauthoryear{Hsieh et~al.}{2015}]{hsieh2015pu}
	\begin{bchapter}
		\bauthor{\bsnm{Hsieh}, \binits{C.-J.}},
		\bauthor{\bsnm{Natarajan}, \binits{N.}},
		\bauthor{\bsnm{Dhillon}, \binits{I.}}:
		\bctitle{Pu learning for matrix completion}.
		In: \bbtitle{International Conference on Machine Learning},
		pp. \bfpage{2445}--\blpage{2453}
		(\byear{2015}).
		\bcomment{PMLR}
	\end{bchapter}
	\endbibitem
	
	%%% 30
	\bibitem[\protect\citeauthoryear{Herbster et~al.}{2016}]{herbster2016mistake}
	\begin{botherref}
		\oauthor{\bsnm{Herbster}, \binits{M.}},
		\oauthor{\bsnm{Pasteris}, \binits{S.}},
		\oauthor{\bsnm{Pontil}, \binits{M.}}:
		Mistake bounds for binary matrix completion.
		Advances in Neural Information Processing Systems
		\textbf{29}
		(2016)
	\end{botherref}
	\endbibitem
	
	%%% 31
	\bibitem[\protect\citeauthoryear{Hammer et~al.}{2023}]{hammer2023approximate}
	\begin{botherref}
		\oauthor{\bsnm{Hammer}, \binits{H.L.}},
		\oauthor{\bsnm{Riegler}, \binits{M.A.}},
		\oauthor{\bsnm{Tjelmeland}, \binits{H.}}:
		Approximate bayesian inference based on expected evaluation.
		Bayesian Analysis,
		1--22
		(2023)
	\end{botherref}
	\endbibitem
	
	%%% 32
	\bibitem[\protect\citeauthoryear{Han et~al.}{2014}]{han2014linear}
	\begin{bchapter}
		\bauthor{\bsnm{Han}, \binits{X.}},
		\bauthor{\bsnm{Wu}, \binits{J.}},
		\bauthor{\bsnm{Wang}, \binits{L.}},
		\bauthor{\bsnm{Chen}, \binits{Y.}},
		\bauthor{\bsnm{Senhadji}, \binits{L.}},
		\bauthor{\bsnm{Shu}, \binits{H.}}:
		\bctitle{Linear total variation approximate regularized nuclear norm
			optimization for matrix completion}.
		In: \bbtitle{Abstract and Applied Analysis},
		vol. \bseriesno{2014}
		(\byear{2014}).
		\bcomment{Hindawi}
	\end{bchapter}
	\endbibitem
	
	%%% 33
	\bibitem[\protect\citeauthoryear{Ji et~al.}{2010}]{ji2010robust}
	\begin{bchapter}
		\bauthor{\bsnm{Ji}, \binits{H.}},
		\bauthor{\bsnm{Liu}, \binits{C.}},
		\bauthor{\bsnm{Shen}, \binits{Z.}},
		\bauthor{\bsnm{Xu}, \binits{Y.}}:
		\bctitle{Robust video denoising using low rank matrix completion}.
		In: \bbtitle{2010 IEEE Computer Society Conference on Computer Vision and
			Pattern Recognition},
		pp. \bfpage{1791}--\blpage{1798}
		(\byear{2010}).
		\bcomment{IEEE}
	\end{bchapter}
	\endbibitem
	
	%%% 34
	\bibitem[\protect\citeauthoryear{Jiang et~al.}{2016}]{jiang2016sparrec}
	\begin{barticle}
		\bauthor{\bsnm{Jiang}, \binits{B.}},
		\bauthor{\bsnm{Ma}, \binits{S.}},
		\bauthor{\bsnm{Causey}, \binits{J.}},
		\bauthor{\bsnm{Qiao}, \binits{L.}},
		\bauthor{\bsnm{Hardin}, \binits{M.P.}},
		\bauthor{\bsnm{Bitts}, \binits{I.}},
		\bauthor{\bsnm{Johnson}, \binits{D.}},
		\bauthor{\bsnm{Zhang}, \binits{S.}},
		\bauthor{\bsnm{Huang}, \binits{X.}}:
		\batitle{Sparrec: An effective matrix completion framework of missing data
			imputation for gwas}.
		\bjtitle{Scientific reports}
		\bvolume{6}(\bissue{1}),
		\bfpage{35534}
		(\byear{2016})
	\end{barticle}
	\endbibitem
	
	%%% 35
	\bibitem[\protect\citeauthoryear{Jewson and Rossell}{2022}]{jewson2022general}
	\begin{barticle}
		\bauthor{\bsnm{Jewson}, \binits{J.}},
		\bauthor{\bsnm{Rossell}, \binits{D.}}:
		\batitle{General bayesian loss function selection and the use of improper
			models}.
		\bjtitle{Journal of the Royal Statistical Society Series B: Statistical
			Methodology}
		\bvolume{84}(\bissue{5}),
		\bfpage{1640}--\blpage{1665}
		(\byear{2022})
	\end{barticle}
	\endbibitem
	
	%%% 36
	\bibitem[\protect\citeauthoryear{Koren et~al.}{2009}]{koren2009matrix}
	\begin{barticle}
		\bauthor{\bsnm{Koren}, \binits{Y.}},
		\bauthor{\bsnm{Bell}, \binits{R.}},
		\bauthor{\bsnm{Volinsky}, \binits{C.}}:
		\batitle{Matrix factorization techniques for recommender systems}.
		\bjtitle{Computer}
		\bvolume{42}(\bissue{8}),
		\bfpage{30}--\blpage{37}
		(\byear{2009})
	\end{barticle}
	\endbibitem
	
	%%% 37
	\bibitem[\protect\citeauthoryear{Kyung et~al.}{2010}]{kyung2010penalized}
	\begin{barticle}
		\bauthor{\bsnm{Kyung}, \binits{M.}},
		\bauthor{\bsnm{Gill}, \binits{J.}},
		\bauthor{\bsnm{Ghosh}, \binits{M.}},
		\bauthor{\bsnm{Casella}, \binits{G.}}:
		\batitle{Penalized regression, standard errors, and bayesian lassos}.
		\bjtitle{Bayesian Analysis}
		\bvolume{5}(\bissue{2}),
		\bfpage{369}--\blpage{412}
		(\byear{2010})
	\end{barticle}
	\endbibitem
	
	%%% 38
	\bibitem[\protect\citeauthoryear{Knoblauch et~al.}{2022}]{Knoblauch}
	\begin{barticle}
		\bauthor{\bsnm{Knoblauch}, \binits{J.}},
		\bauthor{\bsnm{Jewson}, \binits{J.}},
		\bauthor{\bsnm{Damoulas}, \binits{T.}}:
		\batitle{An optimization-centric view on bayes' rule: Reviewing and
			generalizing variational inference}.
		\bjtitle{Journal of Machine Learning Research}
		\bvolume{23}(\bissue{132}),
		\bfpage{1}--\blpage{109}
		(\byear{2022})
	\end{barticle}
	\endbibitem
	
	%%% 39
	\bibitem[\protect\citeauthoryear{Klopp et~al.}{2015}]{klopp2015adaptive}
	\begin{barticle}
		\bauthor{\bsnm{Klopp}, \binits{O.}},
		\bauthor{\bsnm{Lafond}, \binits{J.}},
		\bauthor{\bsnm{Moulines}, \binits{{\'E}.}},
		\bauthor{\bsnm{Salmon}, \binits{J.}}:
		\batitle{Adaptive multinomial matrix completion}.
		\bjtitle{Electronic Journal of Statistics}
		\bvolume{9},
		\bfpage{2950}--\blpage{2975}
		(\byear{2015})
	\end{barticle}
	\endbibitem
	
	%%% 40
	\bibitem[\protect\citeauthoryear{Klopp}{2014}]{klopp2014noisy}
	\begin{barticle}
		\bauthor{\bsnm{Klopp}, \binits{O.}}:
		\batitle{Noisy low-rank matrix completion with general sampling distribution}.
		\bjtitle{Bernoulli}
		\bvolume{20}(\bissue{1}),
		\bfpage{282}--\blpage{303}
		(\byear{2014})
		\doiurl{10.3150/12-BEJ486}
	\end{barticle}
	\endbibitem
	
	%%% 41
	\bibitem[\protect\citeauthoryear{Lyddon et~al.}{2019}]{lyddon2019general}
	\begin{barticle}
		\bauthor{\bsnm{Lyddon}, \binits{S.P.}},
		\bauthor{\bsnm{Holmes}, \binits{C.}},
		\bauthor{\bsnm{Walker}, \binits{S.}}:
		\batitle{General bayesian updating and the loss-likelihood bootstrap}.
		\bjtitle{Biometrika}
		\bvolume{106}(\bissue{2}),
		\bfpage{465}--\blpage{478}
		(\byear{2019})
	\end{barticle}
	\endbibitem
	
	%%% 42
	\bibitem[\protect\citeauthoryear{Lim and Teh}{2007}]{lim2007variational}
	\begin{barticle}
		\bauthor{\bsnm{Lim}, \binits{Y.J.}},
		\bauthor{\bsnm{Teh}, \binits{Y.W.}}:
		\batitle{Variational bayesian approach to movie rating prediction}.
		\bjtitle{Proceedings of KDD cup and workshop}
		\bvolume{7},
		\bfpage{15}--\blpage{21}
		(\byear{2007})
	\end{barticle}
	\endbibitem
	
	%%% 43
	\bibitem[\protect\citeauthoryear{Mai and Alquier}{2015}]{mai2015}
	\begin{barticle}
		\bauthor{\bsnm{Mai}, \binits{T.T.}},
		\bauthor{\bsnm{Alquier}, \binits{P.}}:
		\batitle{A bayesian approach for noisy matrix completion: Optimal rate under
			general sampling distribution}.
		\bjtitle{Electron. J. Statist.}
		\bvolume{9}(\bissue{1}),
		\bfpage{823}--\blpage{841}
		(\byear{2015})
		\doiurl{10.1214/15-EJS1020}
	\end{barticle}
	\endbibitem
	
	%%% 44
	\bibitem[\protect\citeauthoryear{Mai and Alquier}{2017}]{mai2017pseudo}
	\begin{barticle}
		\bauthor{\bsnm{Mai}, \binits{T.T.}},
		\bauthor{\bsnm{Alquier}, \binits{P.}}:
		\batitle{Pseudo-{B}ayesian quantum tomography with rank-adaptation}.
		\bjtitle{Journal of Statistical Planning and Inference}
		\bvolume{184},
		\bfpage{62}--\blpage{76}
		(\byear{2017})
	\end{barticle}
	\endbibitem
	
	%%% 45
	\bibitem[\protect\citeauthoryear{Mai}{2023a}]{mai2023bilinear}
	\begin{barticle}
		\bauthor{\bsnm{Mai}, \binits{T.T.}}:
		\batitle{From bilinear regression to inductive matrix completion: a
			quasi-bayesian analysis}.
		\bjtitle{Entropy}
		\bvolume{25}(\bissue{2}),
		\bfpage{333}
		(\byear{2023})
	\end{barticle}
	\endbibitem
	
	%%% 46
	\bibitem[\protect\citeauthoryear{Mai}{2023b}]{mai2023reduced}
	\begin{barticle}
		\bauthor{\bsnm{Mai}, \binits{T.T.}}:
		\batitle{A reduced-rank approach to predicting multiple binary responses
			through machine learning}.
		\bjtitle{Statistics and Computing}
		\bvolume{33}(\bissue{6}),
		\bfpage{136}
		(\byear{2023})
	\end{barticle}
	\endbibitem
	
	%%% 47
	\bibitem[\protect\citeauthoryear{Merkle}{2005}]{merkle2005jensen}
	\begin{barticle}
		\bauthor{\bsnm{Merkle}, \binits{M.}}:
		\batitle{Jensen's inequality for medians}.
		\bjtitle{Statistics \& probability letters}
		\bvolume{71}(\bissue{3}),
		\bfpage{277}--\blpage{281}
		(\byear{2005})
	\end{barticle}
	\endbibitem
	
	%%% 48
	\bibitem[\protect\citeauthoryear{Matsubara et~al.}{2022}]{matsubara2022robust}
	\begin{barticle}
		\bauthor{\bsnm{Matsubara}, \binits{T.}},
		\bauthor{\bsnm{Knoblauch}, \binits{J.}},
		\bauthor{\bsnm{Briol}, \binits{F.-X.}},
		\bauthor{\bsnm{Oates}, \binits{C.J.}}:
		\batitle{Robust generalised bayesian inference for intractable likelihoods}.
		\bjtitle{Journal of the Royal Statistical Society Series B: Statistical
			Methodology}
		\bvolume{84}(\bissue{3}),
		\bfpage{997}--\blpage{1022}
		(\byear{2022})
	\end{barticle}
	\endbibitem
	
	%%% 49
	\bibitem[\protect\citeauthoryear{Medina et~al.}{2022}]{medina2022robustness}
	\begin{barticle}
		\bauthor{\bsnm{Medina}, \binits{M.A.}},
		\bauthor{\bsnm{Olea}, \binits{J.L.M.}},
		\bauthor{\bsnm{Rush}, \binits{C.}},
		\bauthor{\bsnm{Velez}, \binits{A.}}:
		\batitle{On the robustness to misspecification of $\alpha$-posteriors and their
			variational approximations}.
		\bjtitle{Journal of Machine Learning Research}
		\bvolume{23}(\bissue{147}),
		\bfpage{1}--\blpage{51}
		(\byear{2022})
	\end{barticle}
	\endbibitem
	
	%%% 50
	\bibitem[\protect\citeauthoryear{Park and Casella}{2008}]{park2008bayesian}
	\begin{barticle}
		\bauthor{\bsnm{Park}, \binits{T.}},
		\bauthor{\bsnm{Casella}, \binits{G.}}:
		\batitle{The bayesian lasso}.
		\bjtitle{Journal of the american statistical association}
		\bvolume{103}(\bissue{482}),
		\bfpage{681}--\blpage{686}
		(\byear{2008})
	\end{barticle}
	\endbibitem
	
	%%% 51
	\bibitem[\protect\citeauthoryear{Recht and R{\'e}}{2013}]{recht2013parallel}
	\begin{barticle}
		\bauthor{\bsnm{Recht}, \binits{B.}},
		\bauthor{\bsnm{R{\'e}}, \binits{C.}}:
		\batitle{Parallel stochastic gradient algorithms for large-scale matrix
			completion}.
		\bjtitle{Math. Program. Comput.}
		\bvolume{5}(\bissue{2}),
		\bfpage{201}--\blpage{226}
		(\byear{2013})
		\doiurl{10.1007/s12532-013-0053-8}
	\end{barticle}
	\endbibitem
	
	%%% 52
	\bibitem[\protect\citeauthoryear{Salakhutdinov and
		Mnih}{2008}]{salakhutdinov2008bayesian}
	\begin{bchapter}
		\bauthor{\bsnm{Salakhutdinov}, \binits{R.}},
		\bauthor{\bsnm{Mnih}, \binits{A.}}:
		\bctitle{Bayesian probabilistic matrix factorization using markov chain monte
			carlo}.
		In: \bbtitle{Proceedings of the 25th International Conference on Machine
			Learning},
		pp. \bfpage{880}--\blpage{887}
		(\byear{2008}).
		\bcomment{ACM}
	\end{bchapter}
	\endbibitem
	
	%%% 53
	\bibitem[\protect\citeauthoryear{Syring and
		Martin}{2019}]{syring2019calibrating}
	\begin{barticle}
		\bauthor{\bsnm{Syring}, \binits{N.}},
		\bauthor{\bsnm{Martin}, \binits{R.}}:
		\batitle{Calibrating general posterior credible regions}.
		\bjtitle{Biometrika}
		\bvolume{106}(\bissue{2}),
		\bfpage{479}--\blpage{486}
		(\byear{2019})
	\end{barticle}
	\endbibitem
	
	%%% 54
	\bibitem[\protect\citeauthoryear{Tsybakov et~al.}{2011}]{tsybakov2011nuclear}
	\begin{barticle}
		\bauthor{\bsnm{Tsybakov}, \binits{A.B.}},
		\bauthor{\bsnm{Koltchinskii}, \binits{V.}},
		\bauthor{\bsnm{Lounici}, \binits{K.}}:
		\batitle{Nuclear-norm penalization and optimal rates for noisy low-rank matrix
			completion}.
		\bjtitle{Annals of Statistics}
		\bvolume{39}(\bissue{5}),
		\bfpage{2302}--\blpage{2329}
		(\byear{2011})
	\end{barticle}
	\endbibitem
	
	%%% 55
	\bibitem[\protect\citeauthoryear{Van~Erven and Harremos}{2014}]{van2014renyi}
	\begin{barticle}
		\bauthor{\bsnm{Van~Erven}, \binits{T.}},
		\bauthor{\bsnm{Harremos}, \binits{P.}}:
		\batitle{R{\'e}nyi divergence and kullback-leibler divergence}.
		\bjtitle{IEEE Transactions on Information Theory}
		\bvolume{60}(\bissue{7}),
		\bfpage{3797}--\blpage{3820}
		(\byear{2014})
	\end{barticle}
	\endbibitem
	
	%%% 56
	\bibitem[\protect\citeauthoryear{Yang et~al.}{2018}]{yang2018fast}
	\begin{barticle}
		\bauthor{\bsnm{Yang}, \binits{L.}},
		\bauthor{\bsnm{Fang}, \binits{J.}},
		\bauthor{\bsnm{Duan}, \binits{H.}},
		\bauthor{\bsnm{Li}, \binits{H.}},
		\bauthor{\bsnm{Zeng}, \binits{B.}}:
		\batitle{Fast low-rank {B}ayesian matrix completion with hierarchical gaussian
			prior models}.
		\bjtitle{IEEE Transactions on Signal Processing}
		\bvolume{66}(\bissue{11}),
		\bfpage{2804}--\blpage{2817}
		(\byear{2018})
	\end{barticle}
	\endbibitem
	
	%%% 57
	\bibitem[\protect\citeauthoryear{Yang et~al.}{2020}]{yang2020alpha}
	\begin{barticle}
		\bauthor{\bsnm{Yang}, \binits{Y.}},
		\bauthor{\bsnm{Pati}, \binits{D.}},
		\bauthor{\bsnm{Bhattacharya}, \binits{A.}}:
		\batitle{$\alpha$-variational inference with statistical guarantees}.
		\bjtitle{Annals of Statistics}
		\bvolume{48}(\bissue{2}),
		\bfpage{886}--\blpage{905}
		(\byear{2020})
	\end{barticle}
	\endbibitem
	
	%%% 58
	\bibitem[\protect\citeauthoryear{Yonekura and
		Sugasawa}{2023}]{yonekura2023adaptation}
	\begin{barticle}
		\bauthor{\bsnm{Yonekura}, \binits{S.}},
		\bauthor{\bsnm{Sugasawa}, \binits{S.}}:
		\batitle{Adaptation of the tuning parameter in general bayesian inference with
			robust divergence}.
		\bjtitle{Statistics and Computing}
		\bvolume{33}(\bissue{2}),
		\bfpage{39}
		(\byear{2023})
	\end{barticle}
	\endbibitem
	
\end{thebibliography}
\end{document}